%% file: main.tex
\documentclass[10pt,twocolumn,letterpaper]{article}

\usepackage[pagenumbers]{cvpr} 

\input{preamble}
\definecolor{cvprblue}{rgb}{0.21,0.49,0.74}
\usepackage[pagebackref,breaklinks,colorlinks,allcolors=cvprblue]{hyperref}
\usepackage{multirow}
\usepackage{makecell}
\usepackage{amsmath}
\DeclareMathOperator*{\argmax}{argmax}

\def\paperID{16480} 
\def\confName{CVPR}
\def\confYear{2026}

\title{Syn-GRPO: Self-Evolving Data Synthesis for MLLM Perception Reasoning}

\author{
{Qihan Huang}\textsuperscript{1,2}, {Haofei Zhang}\textsuperscript{1},
{Rong Wei}\textsuperscript{2}, {Yi Wang}\textsuperscript{1}, {Rui Tang}\textsuperscript{2},\\
{Mingli Song}\textsuperscript{1}, {Jie Song}\textsuperscript{1}\\
    \textsuperscript{1} Zhejiang University,
    \textsuperscript{2} Manycore Tech Inc.\\
    {\tt\small \{qh.huang,haofeizhang,y\_w,brooksong,sjie\}@zju.edu.cn,} \\
    {\tt\small \{guangmo,ati\}@qunhemail.com}
}

\begin{document}
\maketitle

\begin{abstract}
RL~(reinforcement learning) methods~(\eg, GRPO) for MLLM~(Multimodal LLM) perception ability has attracted wide research interest owing to its remarkable generalization ability.
Nevertheless, existing reinforcement learning methods still face the problem of \textbf{low data quality}, where data samples cannot elicit diverse responses from MLLMs, thus restricting the exploration scope for MLLM reinforcement learning.
Some methods attempt to mitigate this problem by imposing constraints on entropy, but none address it at its root.
Therefore, to tackle this problem, this work proposes Syn-GRPO~(\textbf{Syn}thesis-\textbf{GRPO}), which employs an online data generator to synthesize high-quality training data with diverse responses in GRPO training.
Specifically, Syn-GRPO consists of two components: (1) data server; (2) GRPO workflow.
The data server synthesizes new samples from existing ones using an image generation model, featuring a decoupled and asynchronous scheme to achieve high generation efficiency.
The GRPO workflow provides the data server with the new image descriptions, and it leverages a diversity reward to supervise the MLLM to predict image descriptions for synthesizing samples with diverse responses.
Experiment results across three visual perception tasks demonstrate that Syn-GRPO improves the data quality by a large margin, achieving significant superior performance to existing MLLM perception methods, and Syn-GRPO presents promising potential for scaling long-term self-evolving RL.
Our code is available at \textit{~\url{https://github.com/hqhQAQ/Syn-GRPO}}.

\end{abstract}

\section{Introduction \label{sec:intro}}

MLLM~(Multimodal LLM) perception ability has drawn widespread research for its role as the prerequisite for all core functions of MLLMs.
Recently, RL~(reinforcement learning) methods~(\eg, GRPO~\cite{guo2025deepseek, shao2024deepseekmath}) fit well with MLLM perception training, owing to its strong generalization and visual perception tasks’ inherent verifiable labels.
Therefore, numerous reasoning methods~(\eg, VLM-R1~\cite{shen2025vlm}, Visual-RFT~\cite{liu2025visual_rft}, VisionReasoner~\cite{liu2025visionreasoner}) based on GRPO have significantly promoted MLLM perception ability.

\input{figs/intro_diversity}

Despite these developments, existing RL methods for MLLM perception still encounter the problem of \textbf{low data quality}.
Low data quality refers to the issue where data samples fail to stimulate diverse responses from MLLMs, thereby limiting the exploration space for MLLM reinforcement learning, as shown in \autoref{fig:intro_diversity}.
To investigate this problem, this work analyzes the \textit{entropy} and \textit{diversity} during GRPO training for perception reasoning.
Specifically, the entropy reflects the uncertainty of the MLLM's output distribution, and the diversity measures the variety of multiple responses for each sample.
\autoref{fig:intro_entropy_diversity} shows that the entropy and diversity of the MLLM exhibit a trend of rapid decline~(referred to as entropy collapse~\cite{yu2025dapo} and diversity collapse), indicating the low quality of visual perception data and resulting in extremely low training efficiency.
Existing LLM RL methods~(\eg, CLIP-Higher~\cite{yu2025dapo}, CLIP-Cov~\cite{cui2025entropy_mechanism}, KL-Cov~\cite{cui2025entropy_mechanism}, Entropy Adv.~\cite{cheng2025reasoning}) attempt to mitigate this problem from GRPO itself; however, they still cannot overcome the limitations of low-quality data at the root.

Therefore, to break the data limit, this work proposes Syn-GRPO~(\textbf{Syn}thesis-\textbf{GRPO}), which leverages an online data generator to synthesize high-quality training data with diverse responses during GRPO training.
Syn-GRPO aims to synthesize new data to replace the old data when processing each training batch, through a self-evolving data synthesis approach.
To this end, Syn-GRPO consists of two components: \textbf{(1) data server}; \textbf{(2) GRPO workflow}.

The data server synthesizes new image data based on the old image data in the visual perception datasets, featuring \textbf{(1) foreground consistency} and \textbf{(2) decoupled design}.
For the first feature, the data server first adopts a foreground segmentation model to remove irrelevant background from the image, and then uses an outpainting model to generate a new background.
The foreground consistency preserves the invariance of visual perception task labels after generation, ensuring accurate supervisory signals during RL training.
Meanwhile, by preserving the foreground, this method also alleviates the problem of excessive distribution discrepancy between new and old data.
For the second feature, the data server adopts a data generation scheme that is decoupled and asynchronous from the GRPO workflow, connected by a unified API.
This separation enables independent management of the data server and GRPO workflow while preserving efficient communication between them.

The GRPO workflow requires the MLLM to predict two outputs in addition to the original reasoning process and final answer: \textbf{(1) new image description} and \textbf{(2) predicted diversity}.
Specifically, the new image description is the text prompt for the new sample, which will be sent to the data server along with the old image to generate the new image.
The predicted diversity represents the MLLM’s estimation of the response diversity for the input sample.
However, generating images from the predicted image descriptions does not necessarily yield highly diverse responses.
To tackle this problem, the GRPO workflow computes a diversity reward by comparing the predicted diversity with the ground-truth diversity from rollout responses.
In this manner, the diversity reward effectively supervises the MLLM to distinguish whether input samples have diverse responses, thereby enhancing its ability to generate image descriptions for samples with diverse responses.
Additionally, this work identifies a diversity drift problem, where the declining trend of the ground-truth diversity causes a shift in the predicted diversity, and accordingly proposes a diversity smoothing strategy to mitigate it.

We perform comprehensive experiments to validate the performance of the proposed Syn-GRPO.
Specifically, we apply Syn-GRPO to three types of visual perception tasks: REC~(Referring Expression Comprehension), OVD~(Open-Vocabulary Object Detection), and ISR~(Indoor Scene Refinement).
The experiment results demonstrate that Syn-GRPO improves the training efficiency of original GRPO by a large margin, achieving significantly superior performance to existing MLLM perception methods.
Furthermore, we reveal two intriguing phenomena: (1) as the size of the original training dataset increases, Syn-GRPO exhibits a stable performance improvement trend; (2) as training progresses through iterations, the generated data shows an increasingly complex trend.
These phenomena indicate that Syn-GRPO, a self-evolving data synthesis approach, has the potential to achieve long-term scalability in RL.

To sum up, the main contributions of this work can be summarized as follows:

$\bullet$ We identify and analyze the problem of low data quality in RL training for MLLM perception.

$\bullet$ We propose Syn-GRPO to tackle this problem, with a data server to efficiently synthesize data in an asynchronous manner, and a GRPO workflow to achieve self-evolving data synthesis through a proposed diversity reward.

$\bullet$ Experiment results across three visual perception tasks demonstrate that Syn-GRPO significantly outperforms existing MLLM perception methods, and Syn-GRPO exhibits promising potential for long-horizon scalability in RL.

\input{figs/intro_entropy_diversity}
\input{figs/framework}

\section{Related Work}

\noindent \textbf{Reinforcement Learning for MLLM Perception.}
After the emergence of DeepSeek-R1, researchers discovered that reinforcement learning~(\eg, GRPO) can significantly improve the reasoning abilities of LLMs and MLLMs by triggering chain-of-thought~(CoT) reasoning, with many studies applying this method to various domains of LLMs and MLLMs.
Among these, RL is particularly well-suited for MLLM perception tasks, as such tasks typically come with verifiable labels~(\eg, bounding box annotations).
Specifically, VLM-R1~\cite{shen2025vlm} and Visual-RFT~\cite{liu2025visual_rft} first demonstrate that GRPO significantly outperforms SFT on visual perception tasks by directly extending GRPO through carefully designed task-specific rewards.
Built upon GRPO, SATORI-R1~\cite{shen2025satori}~(three-stage decomposition), Visionary-R1~\cite{xia2025visionary}~(caption-reason-answer format), UniVG-R1~\cite{bai2025univg}~(difficulty-aware weight adjustment), VisionReasoner~\cite{liu2025visionreasoner}~(unified framework with non-repeat reward), and Rex-Thinker~\cite{jiang2025rex}~(HumanRef-CoT dataset) improve MLLM perception by addressing distinct issues and enhancing framework \& structured reasoning.
However, these methods still suffer from low data quality, resulting in suboptimal performance.

\vspace{0.5em}
\noindent \textbf{Entropy Mechanism of LLM Reinforcement Learning.}
Recently, the entropy mechanism has become a key research focus in LLM RL, because of its close connection with the exploration space of LLM RL.
DAPO~\cite{yu2025dapo} first identified that during GRPO training, the entropy of the model's predicted tokens rapidly decreases.
After this, CLIP-Higher~\cite{yu2025dapo}~(higher GRPO sampling ratio clipping threshold), Clip-Cov~\cite{cui2025entropy_mechanism}~(clip high-covariance tokens), KL-Cov~\cite{cui2025entropy_mechanism}~(KL penalty on high-covariance tokens), and Entropy Adv.~\cite{cheng2025reasoning}~(entropy-included GRPO advantage) attempt to address this issue by imposing constraints on entropy.
Despite their advances, these methods remain fundamentally constrained by data quality and exhibit degraded performance on visual perception tasks where data samples follow more uniform formats.

\vspace{0.5em}
\noindent \textbf{Data Synthesis for LLM Reinforcement Learning.}
Synthetic data holds great potential to address the problem of missing data, and data synthesis methods have emerged for LLM RL.
Specifically, PROMPTCOT~\cite{zhao2025promptcot}~(emulate expert problem designers for math problem synthesis), Genetic-Instruct~\cite{majumdar2025genetic}~(Instructor \& Coder \& Judge-LLM collaborative instruction-code synthesis), METASYNTH~\cite{riaz2025metasynth}~(meta-prompting with multiple-LLM agent collaboration), and TaskCraft~\cite{shi2025taskcraft}~(automated workflow for multi-tool verifiable agent tasks) advance data synthesis via distinct collaborative or workflow-driven strategies.
Absolute Zero~\cite{zhao2025absolute} and R-Zero~\cite{huang2025r_zero} propose an additional generation scheme to synthesize data along with the GRPO training.
However, these data synthesis methods are either inapplicable to visual perception tasks or offline-generated, lacking self-evolution and failing to adaptively synthesize data.

\section{Method \label{sec:method}}

\subsection{Preliminaries \label{sec:method-preliminaries}}

\noindent \textbf{Supervised Fine-tuning~(SFT).}
SFT trains the LLM on curated query-output pairs to improve its instruction-following ability.
The aim of SFT is to maximize the following objective:

\vspace{-0.5em}
\begin{equation}
\label{equa:sft_loss}
    \mathcal{J}_{\rm SFT}(\theta) \!=\! \mathbb{E}[q, o \! \sim \! P(Q,O)] \! \left( \! \frac{1}{|o|} \sum_{t=1}^{|o|} \log \pi_\theta(o_t|q, o_{<t}) \! \right) \!,
\nonumber
\end{equation}

where $q, o$ is the query-output pair sampled from the SFT dataset $P(Q,O)$, $\theta$ denotes the model parameters, $\pi_\theta(o_t|q, o_{<t})$ represents the logit of the model predicting the next token $o_t$ from $q$ and previous tokens $o_{<t}$.

\vspace{0.5em}
\noindent \textbf{Group Relative Policy Optimization~(GRPO).}
GRPO calculates the advantage $\hat{A}_{i,t}$ for the $t$-th token using the average reward of multiple sampled outputs, eliminating the additional value function $V_{\psi}$ in PPO.
Specifically, GRPO samples a group of $G$ outputs $\{ o_1, o_2, ..., o_G \}$ from the old model $\pi_{\theta_{\rm old}}$, then optimizes the model by maximizing the following objective~(\textbf{min \& clip operations} omitted here):

\vspace{-0.5em}
{\small
\begin{align}
& \mathcal{J}_{\rm GRPO}(\theta) = \mathbb{E}[q \sim P(Q), \{o_i\}_{i=1}^G \sim \pi_{\theta_{\rm old}}(O|q)] \nonumber \\
& \frac{1}{G} \sum_{i=1}^G \frac{1}{|o_i|} \sum_{t=1}^{|o_i|} \left\{ \frac{\pi_\theta(o_{i,t}|q,o_{i,<t})}{\pi_{\theta_{\rm old}}(o_{i,t}|q,o_{i,<t})} \hat{A}_{i,t} - \beta D_{\rm KL}[\pi_\theta||\pi_{\rm ref}] \right\}, \nonumber
\end{align}
}

where $D_{\rm KL}[\pi_\theta||\pi_{\rm ref}]$ serves as a regularization term that prevents the new model $\pi_\theta$ from deviating too far from the original model $\pi_{\rm ref}$~(the model before training).
As an ORM method, GRPO provides the reward $r_i$ at the end of each output $o_i$~($r_i = 1$ if the reasoning result is correct, otherwise $r_i = 0$), and sets the advantage $\hat{A}_{i,t}$ of all tokens in $o_i$ as the normalized reward~($\mathbf{r} = \{ r_1, r_2, ..., r_G \}$, $\text{mean}(\cdot)$ denotes the average, and $\text{std}(\cdot)$ denotes the standard deviation):

\vspace{-0.5em}
\begin{equation}
\label{equa:advantage}
    \hat{A}_{i,t} = \tilde{r}_i = \frac{r_i - \text{mean}(\mathbf{r})}{\text{std}(\mathbf{r})}.
\end{equation}

\noindent \textbf{Task-Specific Rewards.}
This work applies Syn-GRPO to three types of visual perception tasks~(REC, OVD, and ISR)~(\autoref{fig:method_tasks}), each with a task-specific accuracy reward.

\textbf{REC}~(Referring Expression Comprehension) aims to localize the object in an image from a given referring expression.
Following VLM-R1~\cite{shen2025vlm}, denote $q, o$ as the query-output pair, $b^{\rm gt}$ as the ground-truth bounding box, $f_{\rm rec}(o)$ as the predicted bounding box extracted from $o$, and ${\rm IoU}(\cdot)$ as the intersection-over-union metric, then the accuracy reward $\mathbf{R}_{\rm acc}^{\rm rec}(o)$ for REC is calculated as:

\vspace{-0.5em}
\begin{equation}
\label{equa:rec_reward}
    \mathbf{R}_{\rm acc}^{\rm rec}(o) = {\rm IoU}(b^{\rm gt}, f_{\rm rec}(o)).
\end{equation}

\textbf{OVD}~(Open-Vocabulary Object Detection) aims to detect the objects in the image and output the corresponding bounding boxes and class labels.
Following VLM-R1~\cite{shen2025vlm}, denote $\hat{b}^{\rm gt}=\{ (b_i^{\rm gt}, c_i^{\rm gt})\}_{i=1}^{N^{\rm gt}}$ as the list of $N^{\rm gt}$ ground-truth bounding-boxes and class labels, $f_{\rm ovd}(o)=\{ (b_i, c_i)\}_{i=1}^{N^{\rm pred}}$ as the list of $N^{\rm pred}$ predicted bounding-boxes and class labels, and ${\rm mAP}(\cdot)$ as the mean average precision metric, then the accuracy reward $\mathbf{R}_{\rm acc}^{\rm ovd}(o)$ for OVD is calculated as~($s_{ovd}$ aims to penalizes redundant predictions):

\vspace{-0.5em}
\begin{equation}
\left\{
\begin{aligned}
\mathbf{R}_{\rm acc}^{\rm ovd}(o) &= s_{\rm ovd} \cdot \text{mAP}(\hat{b}^{\rm gt}, f_{\rm ovd}(o)). \\
s_{\rm ovd} &= \min\biggl(1, \frac{N^{\rm gt}}{N^{\rm pred}}\biggr).
\end{aligned}
\right.
\label{equa:ovd_reward}
\end{equation}

\textbf{ISR}~(Indoor Scene Refinement) refines the indoor scene based on top-view rendering images to enhance its aesthetic quality.
This work decomposes ISR into two stages: perception and refinement, with Syn-GRPO applied in the first stage.
Specifically, the perception stage equals OVD, with ISR’s accuracy reward $\mathbf{R}_{\rm acc}^{\rm isr}(o)$ identical to $\mathbf{R}_{\rm acc}^{\rm ovd}(o)$.
More details about ISR are in {\color{red} \S2.2} of the appendix.

\subsection{Data Quality Analysis \label{sec:method-analysis}}

Although current RL methods have significantly enhanced MLLM perception ability, they suffer from the problem of \textbf{low data quality}.
Low data quality refers to data samples failing to elicit diverse responses from MLLMs, thereby constraining the exploration space in MLLM RL.
To investigate this problem, this work examines the \textbf{entropy} and \textbf{diversity} of MLLM during GRPO training.

\textbf{Entropy} quantifies the predictability or randomness inherent in the tokens predicted by the LLM.
Following Clip-Cov \& KL-Cov~\cite{cui2025entropy_mechanism}, the entropy $\mathcal{H}(q)$ corresponding to sample $q$ is calculated as the average of the entropy of each predicted token~(note that $\pi_\theta$ denotes the model):

\vspace{-0.5em}
{\small
\begin{align}
\mathcal{H}(q) &= -\mathbb{E}_{\{o_i\}_{i=1}^G \sim \pi_{\theta}(O|q)} \left[ \log \pi_\theta(o_{i,t} | q, o_{i,<t}) \right] \nonumber \\
&= - \frac{1}{G} \sum_{i=1}^{G} \frac{1}{|o_i|} \sum_{t=1}^{|o_i|} \mathbb{E}_{o_{i,t} \sim \pi_\theta(O|q)} \left[ \log \pi_\theta(o_{i,t} | q, o_{i,<t}) \right]. \nonumber
\end{align}
}

\textbf{Diversity} measures the extent to which an MLLM generates diverse responses for each sample, quantified by the variance of the accuracy rewards $\mathbf{R}_{\rm acc}(o_i)$ across $G$ responses.
Let $\mathrm{var}(\cdot)$ denote the variance, then the diversity $\mathcal{V}(q)$ corresponding to sample $q$ is calculated as:

\begin{equation}
\mathcal{V}(q) = \mathrm{var}\left( \left\{ \mathbf{R}_{\rm acc}(o_i) \right\}_{i=1}^{G} \right).
\label{equa:diversity}
\end{equation}

In practice, each metric is averaged over the samples in the training batch.
As shown in \autoref{fig:intro_entropy_diversity}, MLLM exhibits entropy collapse and diversity collapse in the visual perception task~(REC), characterized by a rapid decline in both entropy and diversity during GRPO training.
Entropy collapse and diversity collapse indicate that the current visual perception datasets are low-quality, exhibiting overly uniform formatting, which constrains the RL exploration space.  

\input{figs/method_tasks}

\subsection{Syn-GRPO \label{sec:method-syn-grpo}}

To tackle the low-quality data problem, this work proposes Syn-GRPO~(\textbf{Syn}thesis-\textbf{GRPO}), which utilizes an online data generator to adaptively synthesize high-quality training data with diverse responses along with the GRPO training.    
As shown in \autoref{fig:framework}, Syn-GRPO consists of two components: (1) data server; (2) GRPO workflow.

\subsubsection{Data Server}

Previous data synthesis methods use advanced LLMs~(\eg, GPT) to generate new text responses from existing images without modifying the images.
The proposed data server, the first to overcome image-modality data limitations, leverages an image-outpainting model to create new images via original images and new image descriptions.
To accommodate visual perception tasks and enhance training efficiency, this data server incorporates two key features: \textbf{(1) foreground consistency} and \textbf{(2) decoupled design}.

\textbf{Foreground consistency.}
In data synthesis for RL, it is essential to ensure the accuracy of labels in the synthesized data.
Therefore, to preserve visual perception task labels, the data server masks out image regions beyond the ground-truth bounding boxes and then employs an outpainting model to generate new images according to the new image descriptions.
Besides, the data server also employs a foreground segmentation model to more accurately remove the background.
As shown in \autoref{fig:tasks_generated_images}, the labels (\ie, bounding boxes) for visual perception tasks remain unchanged in the new images, ensuring their correctness.
Furthermore, by modifying the original image instead of generating one from scratch, the data server can reduce discrepancies between new and old data distributions.

\textbf{Decoupled design.}
The data server adopts a modular design decoupled from the GRPO workflow, implemented via Python’s \texttt{HTTPServer} package.
Through a unified communication API, the GRPO workflow requests data generation from the data server with the generation parameters and receives the results upon completion.
The decoupled design enables asynchronous execution of GRPO training and data generation, enhancing training efficiency.
Moreover, this design readily supports future extensions to other forms of data synthesis, promoting our method's generalization.

\subsubsection{GRPO workflow}
The GRPO workflow builds upon the original GRPO with three key modifications: \textbf{(1) New image description}, \textbf{(2) Diversity reward}, and \textbf{(3) Description selection}.

\textbf{New image description.}
In addition to the original reasoning process and final answer, the GRPO workflow prompts the MLLM to predict a text description for generating a new high-quality image~(with diverse responses) from the input image.
Specifically, during GRPO training, the MLLM generates $G$ responses $\{ o_i \}_{i=1}^{G}$ for each sample $q$, with each $o_i$ containing a distinct new image description $d_i$ to form the set $\{ d_i \}_{i=1}^{G}$.

\textbf{Diversity reward.}
However, the generated samples from the predicted image descriptions $\{ d_i \}_{i=1}^{G}$ do not ensure highly diverse responses.
Thus, this work proposes a diversity reward to supervise the MLLM in distinguishing whether input samples elicit diverse responses, thereby \textbf{encouraging it to learn image descriptions associated with such diversity}.
To this end, the diversity reward is calculated by comparing the predicted diversity with the ground-truth diversity.
Specifically, the predicted diversity $v_i$ of each response $o_i$ is generated by the MLLM and lies in the range $[0, 1]$.
Next, the ground-truth diversity $\mathcal{V}(q)$ for each sample $q$ is calculated as the variance of accuracy rewards across $G$ responses~(\autoref{equa:diversity}).
Besides, $\mathcal{V}(q)$ is normalized to the range of $[0, 1]$ for the convenience of comparison, using its original upper bound $\frac{1}{4}$ as a reference~(see {\color{red} \S1} of the appendix).
Consequently, the diversity reward $\mathbf{R}_{\rm diversity}(o_i)$ of the $i$-th response $o_i$ is calculated as:

\begin{equation}
\mathbf{R}_{\rm diversity}(o_i) = 1 - |v_i - \mathcal{V}(q) |.
\label{equa:original_diversity_reward}
\end{equation}

Moreover, this work identifies a \textit{diversity drift} problem:
the ground-truth diversity of the model evolves with training; thus, the predicted diversity of the \textbf{current model} is generated for the \textbf{prior model} as it is supervised by the ground-truth diversity of the \textbf{prior model}.
Given the overall declining trend in diversity~(\autoref{fig:intro_entropy_diversity}), this work proposes a \textbf{diversity smoothing method} that mitigates this issue by calibrating the diversity of each training batch using an exponential moving average of historical diversity.
Specifically, let $\mathcal{B}_k$ denote the training batch at the $k$-th step, $\beta \in [0, 1]$ denote the smooth weight, $\mathcal{V}_k^{\rm batch\_avg}$ denote the average diversity of $\mathcal{B}_k$, $\mathcal{V}_k^{\rm global\_avg}$ denote the moving average of diversity at the $k$-th step, then the smoothed ground-truth diversity $\tilde{\mathcal{V}}(q)$ for sample $q$ at the $k$-th step is calculated as~(note that ${\rm clip}(\cdot)$ clamps a value to a range):

\vspace{-0.5em}
\begin{equation}
\left\{
\begin{aligned}
& \mathcal{V}_k^{\rm batch\_avg} = \frac{1}{|\mathcal{B}_k|} \sum\limits_{q'\in \mathcal{B}_k} \mathcal{V}(q'). \\
& \mathcal{V}_k^{\rm global\_avg} = \beta \cdot \mathcal{V}_{k-1}^{\rm global\_avg} + (1 - \beta) \cdot \mathcal{V}_k^{\rm batch\_avg}. \nonumber \\
& \tilde{\mathcal{V}}(q) = {\rm clip} \left(\mathcal{V}(q) \cdot \frac{\mathcal{V}_k^{\rm global\_avg}}{\mathcal{V}_k^{\rm batch\_avg}}, 0, 1 \right). \nonumber
\end{aligned}
\right.
\label{equa:smooth_diversity}
\end{equation}

Note that $\mathcal{V}_1^{\rm global\_avg}$, the $\mathcal{V}_k^{\rm global\_avg}$ at the first step, is equal to $\mathcal{V}_1^{\rm batch\_avg}$.
The diversity smoothing method aligns the current model’s diversity distribution with the previous one through the ratio of $\mathcal{V}_k^{\rm global\_avg}$ and $\mathcal{V}_k^{\rm batch\_avg}$, enhancing reward training accuracy and stability.
Finally, the updated diversity reward $\mathbf{R}_{\rm diversity}(o_i)$ is calculated as in \autoref{equa:diversity_reward}, which also lies in the range $[0, 1]$.

\vspace{-0.5em}
\begin{equation}
\mathbf{R}_{\rm diversity}(o_i) = 1 - |v_i - \tilde{\mathcal{V}}(q) |.
\label{equa:diversity_reward}
\end{equation}

Therefore, the final reward $r_i$ for the $i$-th response $o_i$ is calculated as the summation of its accuracy reward $\mathbf{R}_{\rm acc}(o_i)$, format reward $\mathbf{R}_{\rm format}(o_i)$, and diversity reward $\mathbf{R}_{\rm diversity}(o_i)$.
Note that $\mathbf{R}_{\rm format}(o_i)$ supervises $o_i$ to incorporate the reasoning process, predicted diversity, new image description, and the final answer.

\vspace{-0.5em}
\begin{equation}
r_i = \mathbf{R}_{\rm acc}(o_i) + \mathbf{R}_{\rm format}(o_i) + \mathbf{R}_{\rm diversity}(o_i).
\label{equa:final_reward}
\end{equation}

\input{tables/benchmark_rec}
\input{tables/benchmark_ovd}

\textbf{Description selection.}
For the $G$ new image descriptions $\{ d_i \}_{i=1}^{G}$ derived from the MLLM-generated responses $\{ o_i \}_{i=1}^{G}$, the GRPO workflow requires selecting one for transmission to the data server.
To this end, this work selects the image description with the highest $\mathbf{R}_{\rm diversity}(o_i)$, as this response more accurately assesses the diversity of answers for the input $q$, leading to a more accurate image description for the new image with diverse responses.
Specifically, the selected index $i^{*}$ is calculated as in \autoref{equa:selected_i}, and the corresponding selected description is $d_{i^{*}}$.

\vspace{-0.5em}
\begin{equation}
i^{*} = \argmax_{i \in \{ 1,2,...,G \}} \mathbf{R}_{\rm diversity}(o_i).
\label{equa:selected_i}
\end{equation}

\section{Experiments \label{sec:experiments}}

\noindent \textbf{Implementation details.}
Following existing MLLM perception methods~(\eg, VLM-R1~\cite{shen2025vlm}, Visionary-R1~\cite{xia2025visionary}, and SATORI-R1~\cite{shen2025satori}), we conduct the main experiments using Qwen2.5-VL-3B~\cite{bai2025qwen2} and Qwen2.5-VL-7B~\cite{bai2025qwen2} as base models.
This work implements Syn-GRPO on the verl framework~\cite{sheng2024hybridflow}, leveraging the vLLM engine~\cite{kwon2023efficient} for RL rollout acceleration and FSDP~\cite{zhao2023fsdp} for distributed training.
In the data server, the image-outpainting model is a ControlNet~\cite{zhang2023controlnet} based on the SDXL~\cite{podell2023sdxl} model, and the foreground segmentation model is BEN2~(Background Erase Network)~\cite{meyer2025ben}.
For all three types of visual perception tasks, we adopt the AdamW optimizer~\cite{ilya2019adamw} to train the model for 5 epochs, with a learning rate of $1\times10^{-6}$, a batch size of 20, and a rollout number $G$ of 6.
Besides, we allocate 4 GPUs for the GRPO workflow, and 1 GPU for the data server.
The hyper-parameter $\gamma$ in the diversity reward is set to 0.7, as validated by ablation experiments.

\vspace{0.5em}
\noindent \textbf{Training datasets.}
For the REC task, we follow VLM-R1 to use the RefCOCO \& RefCOCO+ \& RefCOCOg datasets~\cite{mao2016generation, yu2016referring} for training.
Specifically, we randomly select 2,000 samples from their training sets and train for 5 epochs~(resulting in 10,000 sample updates), which is comparable to VLM-R1’s total training budget of 9,600 samples.
For the OVD task, we also follow VLM-R1 to use the D${}^3$ dataset~\cite{xie2023d3dataset} for training, and randomly select 2,000 samples likewise.
For the ISR task, we utilize the 3D-FRONT dataset~\cite{fu2021front3d} for training, and randomly select 2,000 samples.

\input{figs/tasks_generated_images}

\vspace{0.5em}
\noindent \textbf{Test datasets.}
For the REC task, following VLM-R1, we evaluate on out-of-domain data~(the test set of LISA-Grounding~\cite{lai2024lisa}) and in-domain data~(validation sets of RefCOCO, RefCOCO+, and RefCOCOg).
For the OVD task, we also follow VLM-R1 to evaluate on the COCO${}_{\rm filter}$ dataset~\cite{lin2014coco}, a filtered subset derived from the COCO validation set.
For the ISR task, we randomly select 500 samples from the 3D-FRONT dataset for evaluation.

\vspace{0.5em}
\noindent \textbf{Evaluation Metric.}
For the REC task, we follow VLM-R1 and calculate the accuracy as the ratio of predicted bounding boxes with IoU $>$ 0.5 relative to the ground-truth box.
For the OVD task and the perception stage of ISR task, we follow VLM-R1 to utilize mAP~(mean Average Precision), GP~(Greedy Precision), and GR~(Greedy Recall) for evaluation.
Specifically, GP is the fraction of predicted boxes that match any ground-truth box, while GR is the reverse.
For the refinement stage of ISR task, we employ $S_{\rm bbox}$ and $S_{\rm rotation}$ to evaluate the refinement quality, assessing object position and rotation, respectively.
More details about GP, GR, $S_{\rm bbox}$, and $S_{\rm rotation}$ are in {\color{red} \S2.3} of the appendix.


\input{tables/benchmark_isr}

\subsection{Comparison Analysis}

\noindent \textbf{REC.}
\autoref{tab:benchmark_rec} demonstrates the performance comparison of different methods on the REC task, evaluated with two base MLLMs: Qwen2.5-VL-3B and Qwen2.5-VL-7B.
Several conclusions can be drawn from \autoref{tab:benchmark_rec}.


(1) In the upper portion of the table, RL methods with entropy constraints~(Entropy Loss, Entropy Adv.) partially mitigate entropy collapse, preserving exploration capacity on low-quality data.
However, they fail to fundamentally resolve the core data quality issue, yielding only marginal improvements over GRPO.

(2) In the lower portion of the table, GRPO underperforms VLM-R1, attributed to its utilization of fewer training samples~(2,000 versus 9,600).

(3) In the lower portion of the table, Syn-GRPO outperforms both standard RL methods, entropy-based methods, and the offline generation method~(using the descriptions from GPT-4o) by synthesizing high-quality data through a self-evolutionary mechanism.

(4) Syn-GRPO exhibits more pronounced performance gains on out-of-domain data~(LISA-Grounding) than on in-domain benchmarks~(RefCOCO, RefCOCO+, and RefCOCOg), as the synthesized data spans a broader, more comprehensive distribution, endowing the MLLM with more robust and comprehensive perception abilities.

\vspace{0.5em}
\noindent \textbf{OVD \& ISR.}
\autoref{tab:benchmark_ovd} and \autoref{tab:benchmark_isr} show that Syn-GRPO also surpasses GRPO in these two visual perception tasks. Moreover, the performance gain is more pronounced on OVD than on ISR, as the training and test datasets in OVD are out-of-domain, whereas those in ISR are in-domain.
Furthermore, experiments in {\color{red} \S2.5} of the appendix demonstrate that Syn-GRPO also enhances the performance of the refinement stage~(\ie, second stage) of ISR.

\input{figs/acc_sizes}
\input{figs/diversity_smooth}

\subsection{Ablation Experiments}

This section presents ablation studies of Syn-GRPO:
(1) Variations in entropy \& diversity;
(2) Effect of data size;
(3) Diversity smoothing;
(4) Asynchronous data synthesis.

\vspace{0.5em}
\noindent \textbf{Variations in entropy \& diversity.}
As shown in \autoref{fig:intro_entropy_diversity}, during the original GRPO training, the entropy and diversity of the MLLM exhibit a rapid decreasing trend.
By contrast, Syn-GRPO sustains high entropy and diversity from the first epoch~(20\% training progress), substantially mitigating the issues of entropy collapse and diversity collapse using the newly generated samples.

\vspace{0.5em}
\noindent \textbf{Effect of data size.}
This experiment varies data size~(400 to 2,000 samples) in the REC task, investigating the impact of data size on model performance.
As shown in \autoref{fig:acc_sizes}, Syn-GRPO performance displays a steady upward trend with increasing data size, indicating a data scaling law and suggesting strong potential for generalization to long-term reinforcement learning.

\vspace{0.5em}
\noindent \textbf{Diversity smoothing.}
This experiment verifies the effect of the diversity smoothing method on the proposed diversity reward.
\autoref{fig:diversity_smooth}~(a) demonstrates that, with diversity smoothing, the diversity reward increases more steadily~(with reduced fluctuations).
Moreover, it also shows that diversity smoothing leads to a more rapid increase in the diversity reward.
\autoref{fig:diversity_smooth}~(b) presents that, with diversity smoothing, the test accuracy increases more rapidly and ultimately reaches a higher plateau.
Overall, these two figures verify that diversity smoothing effectively aligns the current model’s diversity distribution with the previous one, yielding a more accurate diversity reward and enabling stable and precise training.
Furthermore, \autoref{tab:ablation_gamma} shows that performance improves with initial increases in $\gamma$ but degrades when $\gamma$ is excessively large, as it restricts the update of smoothed diversity.

\input{figs/rec_image_trends}
\input{figs/rec_surreal}
\input{tables/ablation_gamma}

\vspace{0.5em}
\noindent \textbf{Asynchronous data synthesis.}
\autoref{tab:training_time} shows that the synchronous data server design nearly doubles training time versus the original GRPO, as it must generate new data for each batch before proceeding.
To tackle this problem, the data server adopts a modular design decoupled from the GRPO workflow, enabling asynchronous data synthesis with negligible overhead, as shown in \autoref{tab:training_time}.

\subsection{Visualization Analysis}

\noindent \textbf{Visualization of generated images.}
\autoref{fig:tasks_generated_images} presents examples of generated images for three types of visual perception tasks.
These visualizations show that the generated images possess high image fidelity and foreground consistency, thereby ensuring the accuracy of labels for the new data and satisfying the requirements for RL training.

\vspace{0.5em}
\noindent \textbf{Trends in generated images.}
\autoref{fig:rec_image_trends} illustrates the generated images from the same original image at various epochs, highlighting their increasing complexity and difficulty, thereby preserving the diversity in reinforcement learning.
This phenomenon demonstrates Syn-GRPO's potential for adapting to long-term scalable RL by generating increasingly challenging images throughout RL training.

\vspace{0.5em}
\noindent \textbf{Surreal generated images.}
We observe an intriguing phenomenon wherein the MLLM occasionally generates surreal textual descriptions for new images, as illustrated in \autoref{fig:rec_surreal}.
Although these images do not exist in reality, they hold potential to benefit RL training by stimulating the MLLM’s reasoning capability through eliciting novel reasoning pathways from new perspectives.

\vspace{0.5em}
\noindent \textbf{Visualization w/ \& w/o Syn-GRPO.}
\autoref{fig:rec_difficult} demonstrates that, in contrast to GRPO, Syn-GRPO accurately localizes a partially occluded recreational vehicle.
This indicates that Syn-GRPO enhances the MLLM's perception ability under challenging conditions such as object occlusion and complex backgrounds, by synthesizing a large volume of high-quality, complex training data.

Furthermore, we have provided more visualization analysis in {\color{red} \S3} of the appendix.

\input{figs/rec_difficult}
\input{tables/training_time}

\section{Conclusion}

In this work, we provide a thorough analysis of the low-quality data problem in RL for MLLM perception, where data samples fail to elicit diverse responses from MLLMs, thereby limiting the exploration scope for RL.
To tackle this problem, we propose Syn-GRPO~(\textbf{Syn}thesis-\textbf{GRPO}), which utilizes an online data generator to synthesize high-quality training data with diverse responses in GRPO training.
Specifically, Syn-GRPO consists of a data server and a GRPO workflow.
The data server leverages an image generation model to synthesize new samples with an asynchronous design.
The GRPO workflow provides new image descriptions to the server and proposes a diversity reward, supervising the MLLM to predict descriptions for high-quality images with diverse responses.
Experimental results across three visual perception tasks indicate that Syn-GRPO substantially enhances data quality, significantly outperforming existing RL methods.
We hope our framework can contribute to the community of MLLM reasoning.


{
    \small
    \bibliographystyle{ieeenat_fullname}
    \bibliography{main}
}


\end{document}

%% file: figs/intro_diversity.tex
\begin{figure}[t]
\centering
    \includegraphics[width=1.0\linewidth]{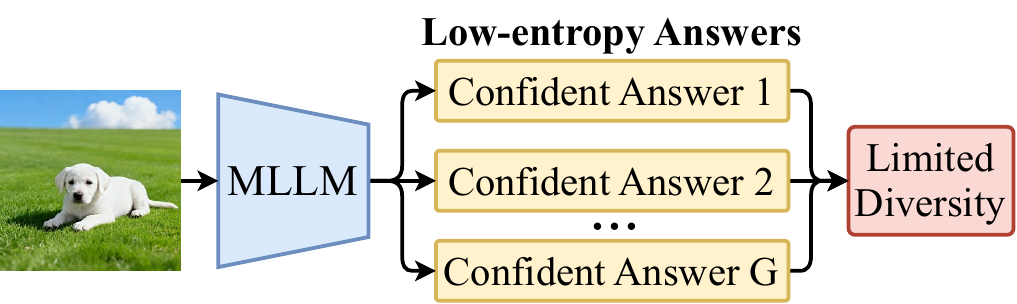}
\caption{In GRPO training, the MLLM generates non-diverse answers with low entropy, restricting the exploration space of RL.}
\vspace{-1em}
\label{fig:intro_diversity}
\end{figure}

%% file: figs/intro_entropy_diversity.tex
\begin{figure}[t]
\centering
    \includegraphics[width=1.0\linewidth]{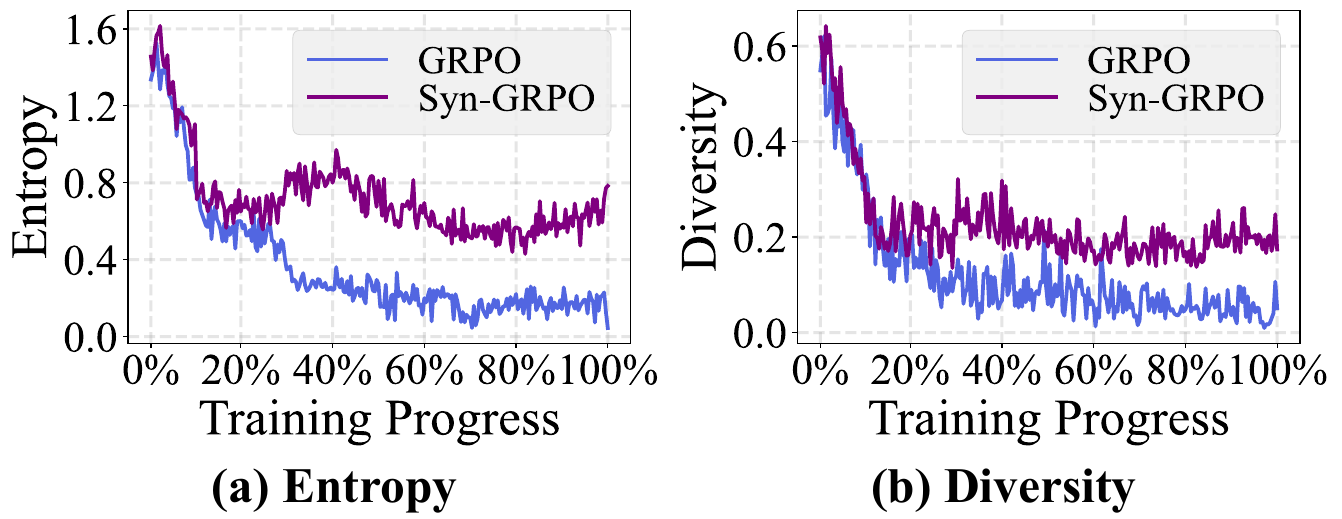}
\caption{Entropy collapse and diversity collapse of GRPO for Qwen2.5-VL-3B on the visual perception task~(REC).
Note that 20\% training progress corresponds to one training epoch.}
\vspace{-1em}
\label{fig:intro_entropy_diversity}
\end{figure}

%% file: figs/framework.tex
\begin{figure*}[t]
\centering
    \includegraphics[width=1.0\linewidth]{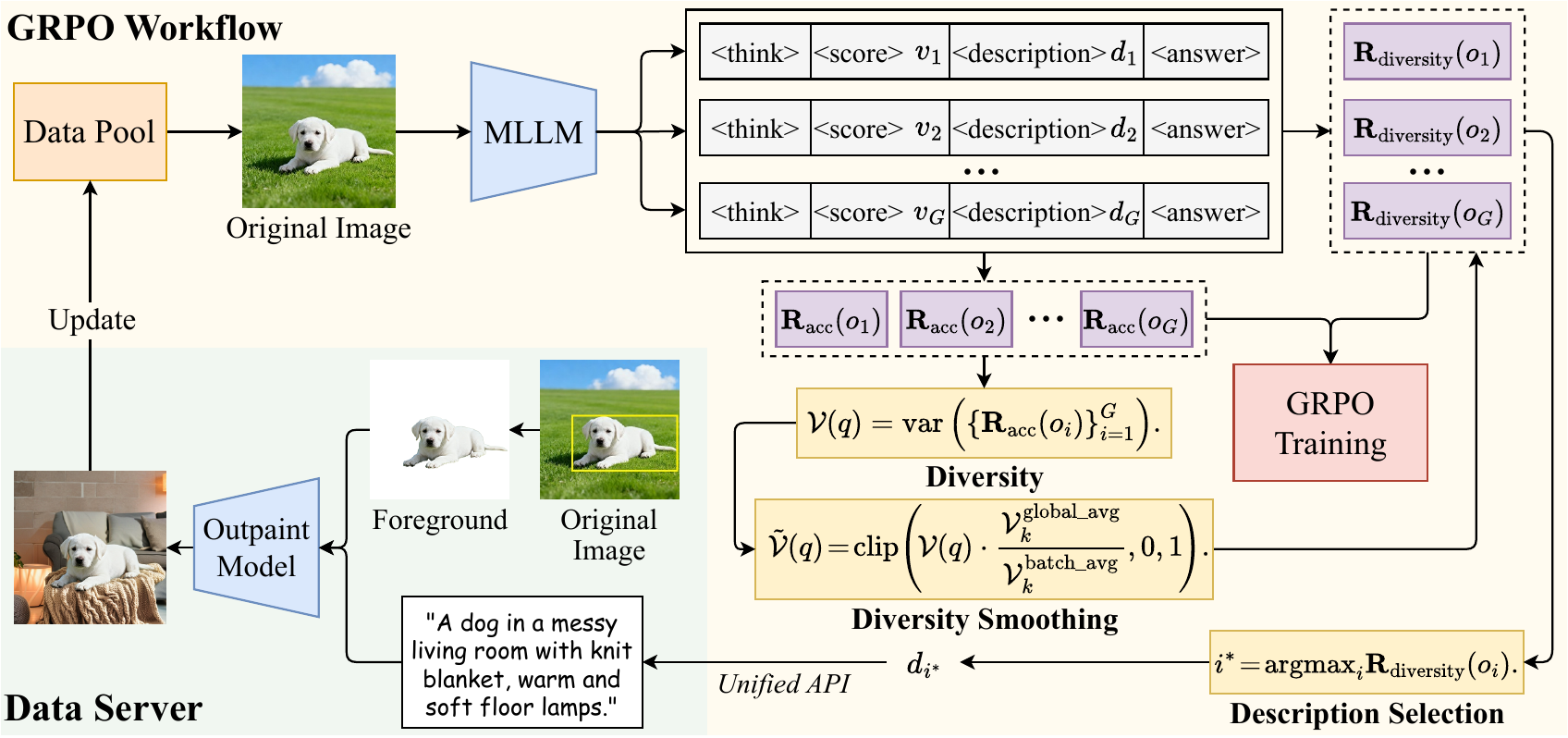}
\caption{Framework of Syn-GRPO. Syn-GRPO consists of a data server and a GRPO workflow.
The data server generates new high-quality images with diverse responses, based on the original images and image descriptions $d_{i^*}$.
The GRPO workflow predicts new image descriptions for the data server, and it employs a diversity reward $\mathbf{R}_{\rm diversity}(o_i)$ to supervise the generation of these descriptions.
}
\vspace{-1em}
\label{fig:framework}
\end{figure*}

%% file: figs/method_tasks.tex
\begin{figure}[t]
\centering
    \includegraphics[width=1.0\linewidth]{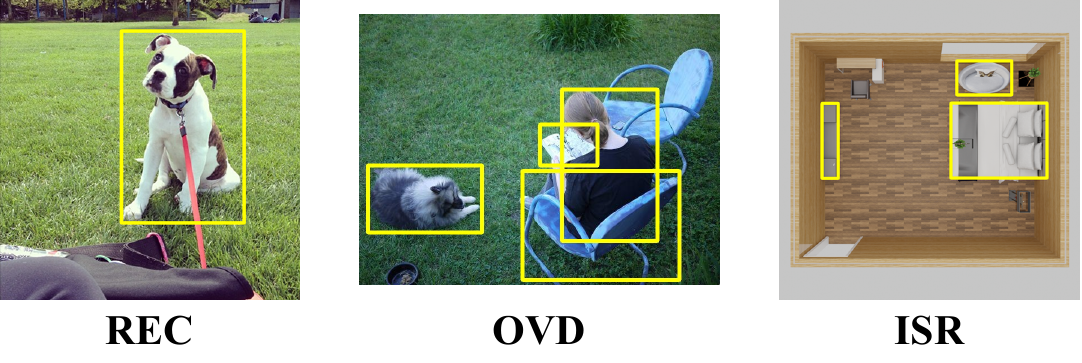}
\caption{Samples of three visual perception tasks~(REC, OVD, and ISR) with the bounding box annotations.}
\vspace{-1em}
\label{fig:method_tasks}
\end{figure}

%% file: tables/benchmark_rec.tex
\begin{table*}
\small
\renewcommand\arraystretch{0.9}
\centering
\setlength{\tabcolsep}{1.63mm}
\begin{tabular}{c *{4}{c}|*{4}c}
  \toprule
  \multirow{2}{*}{\textbf{Method}} 
  & \multicolumn{4}{c|}{\textbf{\textit{Qwen2.5-VL-3B}}} 
  & \multicolumn{4}{c}{\textbf{\textit{Qwen2.5-VL-7B}}} \\
  \cmidrule(lr){2-9}
  & \textbf{\small LISA} & \textbf{\small RefCOCO} & \textbf{\small RefCOCO+} & \textbf{\small RefCOCOg} 
  & \textbf{\small LISA} & \textbf{\small RefCOCO} & \textbf{\small RefCOCO+} & \textbf{\small RefCOCOg} \\
  \midrule
  {\small Original} & 56.51 & 88.70 & 81.95 & 86.05 & 61.34 & 90.00 & 84.20 & 87.20 \\
  {\small Visionary-R1} & 60.80 & 86.85 & 82.65 & 86.50 & N/A & N/A & N/A & N/A \\
  {\small SATORI-R1} & 58.38 & 85.80 & 82.20 & 85.25 & N/A & N/A & N/A & N/A \\
  {\small Rex-Thinker} & N/A & N/A & N/A & N/A & 67.49 & 91.20 & 86.35 & 87.80 \\
  {\small SFT} & 54.82 & 88.70 & 82.25 & 85.95 & 63.27 & 89.10 & 85.95 & 86.65 \\
  {\small VLM-R1} & 63.14 & 90.55 & 84.30 & 87.10 & 66.71 & 92.60 & 89.40 & 89.00 \\
  \midrule
  {\small GRPO} & 62.24 & 89.40 & 84.10 & 86.60 & 65.26 & 91.90 & 89.85 & 87.80 \\
  {\small GRPO + Entropy Loss} & 64.23 & 90.20 & 82.55 & 85.60 & 66.59 & 91.40 & 87.65 & 87.95 \\
  {\small GRPO + Entropy Adv.} & 63.69 & 90.90 & 83.85 & 86.40 & 67.25 & 92.05 & 88.30 & 87.30 \\
  {\small GRPO + Offline Generation} & 64.23 & 91.10 & 83.40 & 86.30 & 66.95 & 91.80 & 89.85 & 88.55 \\
  {\small \bf Syn-GRPO}~(w/o $\mathbf{R}_{\rm diversity}$) & 64.60 & 91.35 & 83.85 & 85.85 & 67.67 & 92.75 & 89.50 & 88.75 \\
  {\small \bf Syn-GRPO} & \textbf{68.28} & \textbf{92.15} & \textbf{85.30} & \textbf{87.45} & \textbf{70.14} & \textbf{93.55} & \textbf{90.65} & \textbf{89.25} \\
  \bottomrule
\end{tabular}
\caption{Experiment results of two MLLMs on the REC task.
The table has two sections: upper for existing models' performance,
lower for existing methods \& our method trained in the same setting. LISA abbreviates LISA-Grounding, and bold font denotes the best result.
}
\vspace{-1em}
\label{tab:benchmark_rec}
\end{table*}

%% file: tables/benchmark_ovd.tex
\begin{table}
\small
\renewcommand\arraystretch{0.9}
\centering
\setlength{\tabcolsep}{2.8mm}{
\begin{tabular}{c|*{3}{c}}
  \toprule

\textbf{Method} & \textbf{\small mAP} & \textbf{\small GP~(IoU=0.5)} & \textbf{\small GR~(IoU=0.5)} \\
\midrule

{\small Original} & 14.20 & 56.06 & 33.79 \\
{\small SFT} & 18.50 & 53.15 & 39.40 \\
{\small VLM-R1} & 21.10 & 67.34 & 43.84 \\
{\small GRPO} & 18.66 & 63.83 & 36.95 \\

\midrule
{\small \bf Syn-GRPO} & \textbf{23.74} & \textbf{71.42} & \textbf{46.44} \\

\bottomrule
\end{tabular}}
\caption{Experiment results on the OVD task.}
\vspace{-1em}
\label{tab:benchmark_ovd}
\end{table}

%% file: figs/tasks_generated_images.tex
\begin{figure}[t]
\centering
    \includegraphics[width=1.0\linewidth]{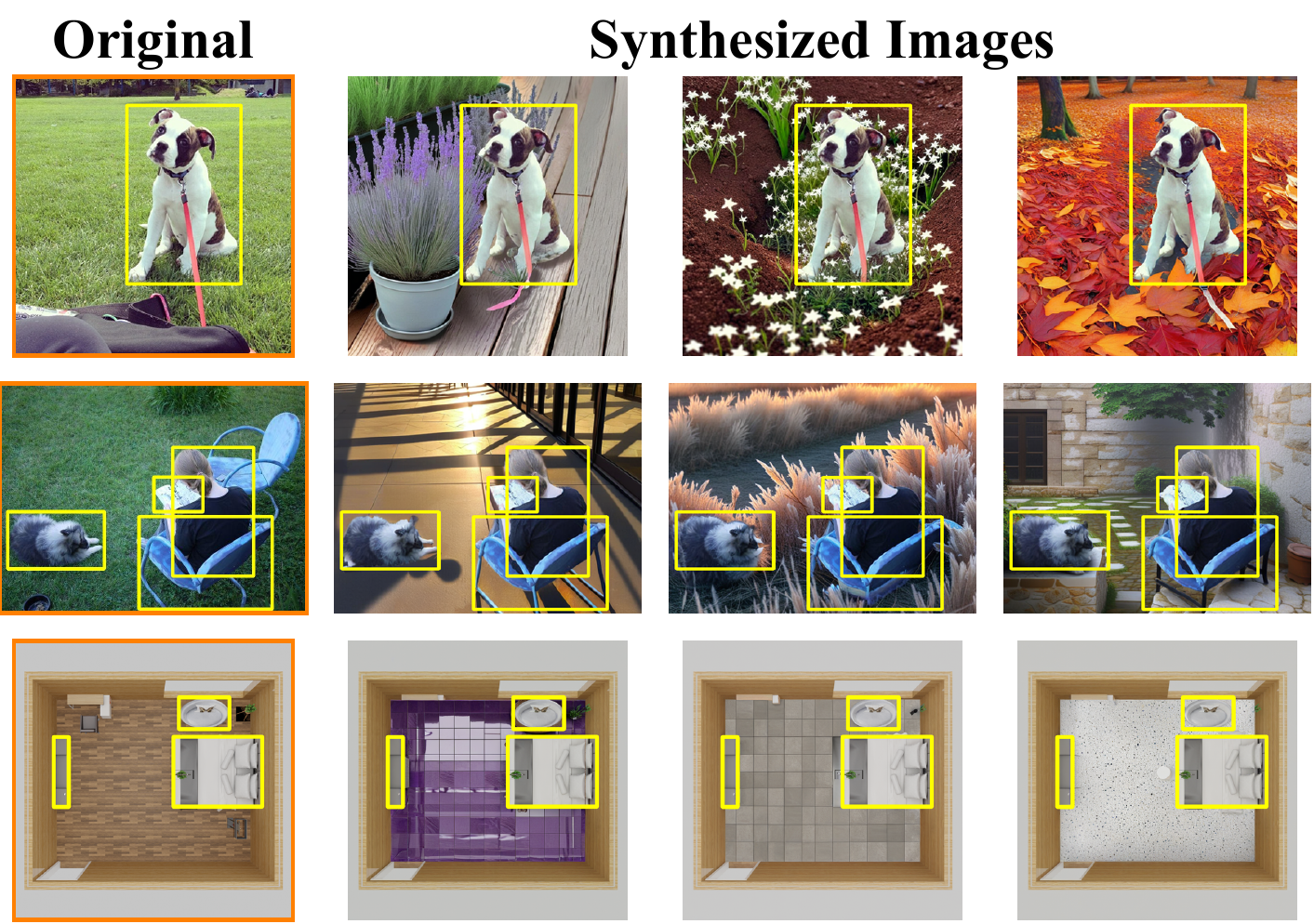}
\caption{Synthesized images of three visual perception tasks.}
\vspace{-1em}
\label{fig:tasks_generated_images}
\end{figure}

%% file: tables/benchmark_isr.tex
\begin{table}
\small
\renewcommand\arraystretch{0.9}
\centering
\setlength{\tabcolsep}{2.8mm}{
\begin{tabular}{c|*{3}{c}}
  \toprule

\textbf{Method} & \textbf{\small mAP} & \textbf{\small GP~(IoU=0.5)} & \textbf{\small GR~(IoU=0.5)} \\
\midrule

{\small Original} & 26.18 & 65.20 & 42.52 \\
{\small SFT} & 28.11 & 69.30 & 44.99 \\
{\small GRPO} & 33.46 & 74.12 & 50.98 \\

\midrule
{\small \bf Syn-GRPO} & \textbf{35.77} & \textbf{76.32} & \textbf{53.52} \\

\bottomrule
\end{tabular}}
\caption{Experiment results on the first stage of ISR task.}
\vspace{-1em}
\label{tab:benchmark_isr}
\end{table}

%% file: figs/acc_sizes.tex
\begin{figure}[t]
\centering
    \includegraphics[width=1.0\linewidth]{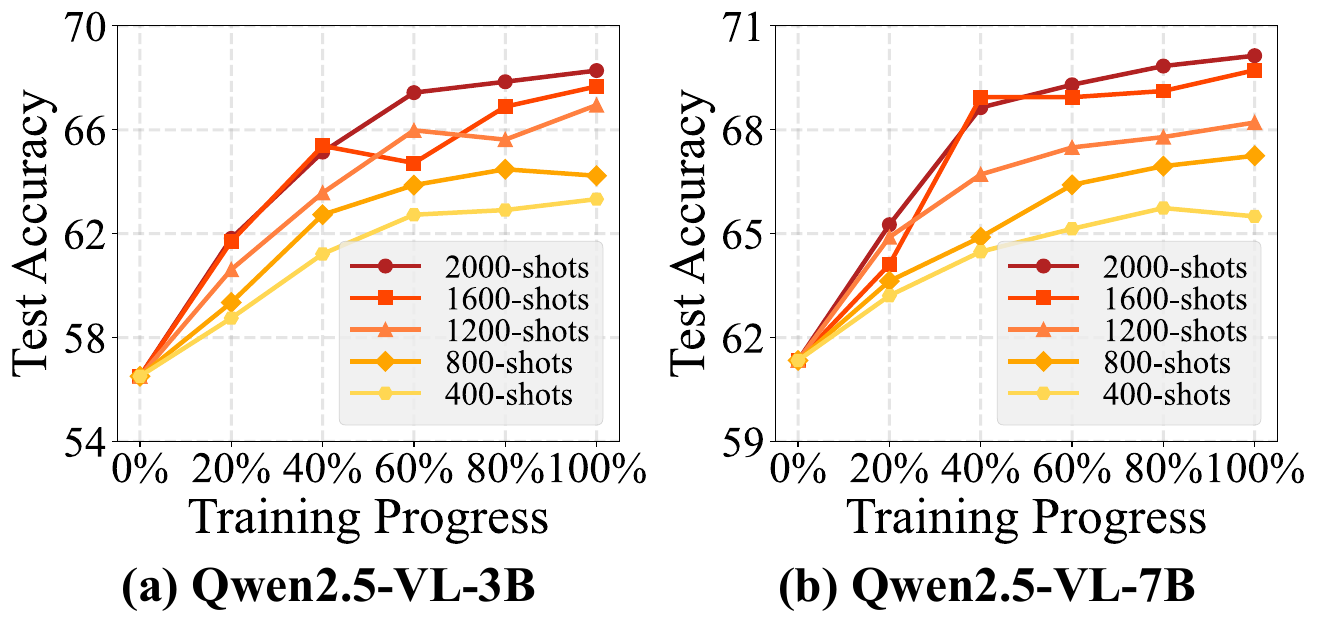}
\caption{Ablation experiments of data size for Syn-GRPO on the LISA-Grounding dataset.}
\vspace{-1em}
\label{fig:acc_sizes}
\end{figure}

%% file: figs/diversity_smooth.tex
\begin{figure}[t]
\centering
    \includegraphics[width=1.0\linewidth]{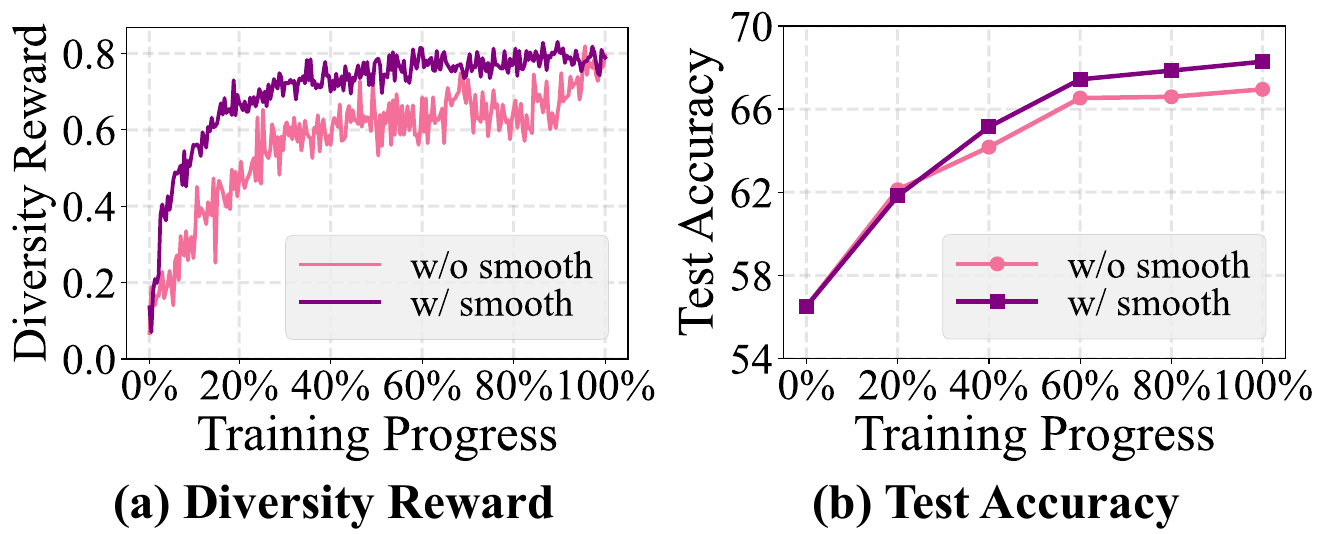}
\caption{Ablation experiments of the diversity smoothing method for Qwen2.5-VL-3B on the LISA-Grounding dataset.}
\vspace{-1em}
\label{fig:diversity_smooth}
\end{figure}

%% file: figs/rec_image_trends.tex
\begin{figure}[t]
\centering
    \includegraphics[width=1.0\linewidth]{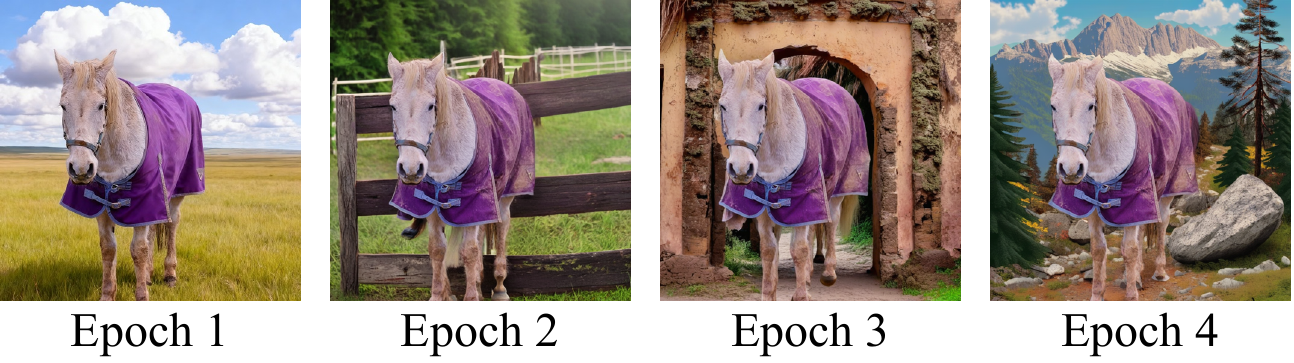}
\vspace{-1.5em}
\caption{Generated images show a growingly complex trend.}
\label{fig:rec_image_trends}
\end{figure}

%% file: figs/rec_surreal.tex
\begin{figure}[t]
\centering
    \includegraphics[width=1.0\linewidth]{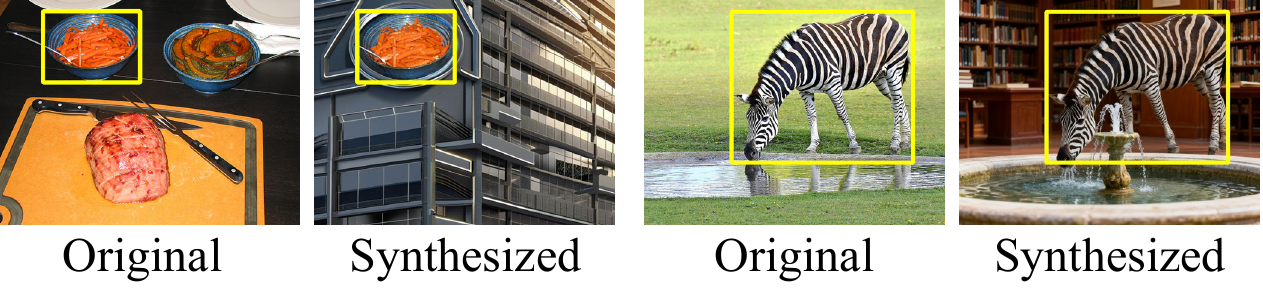}
\vspace{-1.5em}
\caption{Two examples of surreal generated images.}
\vspace{-0.5em}
\label{fig:rec_surreal}
\end{figure}

%% file: tables/ablation_gamma.tex
\begin{table}
\small
\renewcommand\arraystretch{0.9}
\centering
\setlength{\tabcolsep}{2.1mm}{
\begin{tabular}{c|*{5}{c}}
  \toprule

\textbf{Model} & \textbf{\small 0.1} & \textbf{\small 0.3} & \textbf{\small 0.5} & \textbf{\small 0.7$^{\dagger}$} & \textbf{\small 0.9} \\
\midrule

{\small Qwen2.5-VL-3B} & 67.43 & 67.31 & 67.85 & \textbf{68.28} & 67.55 \\
{\small Qwen2.5-VL-7B} & 68.88 & 69.60 & 69.84 & \textbf{70.14} & 69.24 \\

\bottomrule
\end{tabular}}
\caption{Ablation experiments of $\gamma$ for diversity smoothing on the LISA-Grounding dataset. $\dagger$ denotes the selected one.}
\vspace{-1em}
\label{tab:ablation_gamma}
\end{table}

%% file: figs/rec_difficult.tex
\begin{figure}[t]
\centering
    \includegraphics[width=1.0\linewidth]{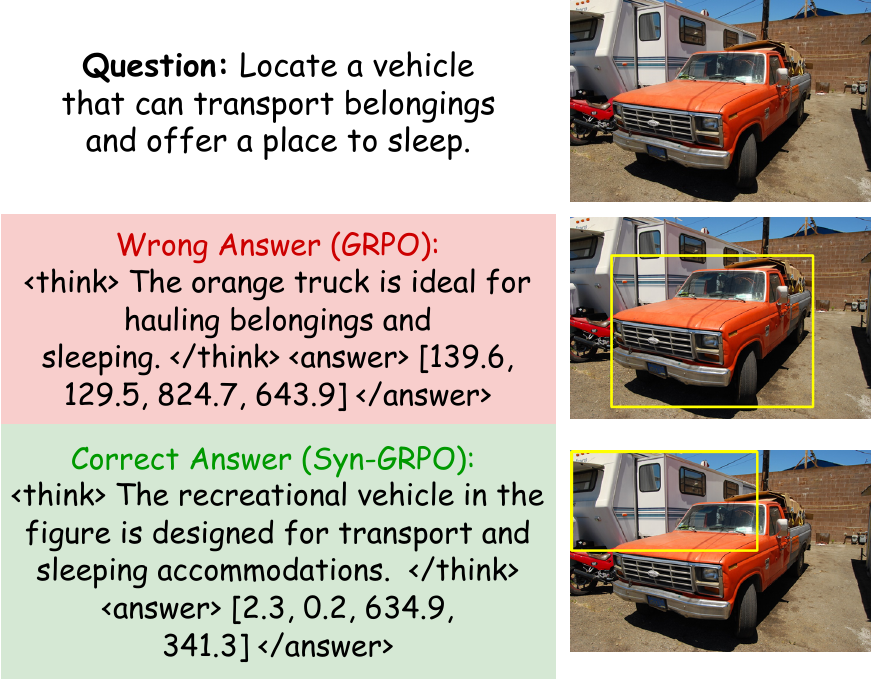}
\caption{Example of Syn-GRPO's outputs.}
\vspace{-1em}
\label{fig:rec_difficult}
\end{figure}

%% file: tables/training_time.tex
\begin{table}
\small
\renewcommand\arraystretch{0.9}
\centering
\setlength{\tabcolsep}{1.05mm}{
\begin{tabular}{c|*{3}{c}}
  \toprule

\textbf{Model} & \textbf{\small GRPO} & \textbf{\small Synchronous} & \textbf{\small Asynchronous} \\
\midrule

{\small Qwen2.5-VL-3B} & {\color{gray} 6.10 Hours} & 13.16 Hours & \textbf{6.45} Hours \\
{\small Qwen2.5-VL-7B} & {\color{gray} 13.66 Hours} & 28.47 Hours & \textbf{13.93} Hours \\

\bottomrule
\end{tabular}}
\caption{Training time of original GRPO on the REC task, Syn-GRPO~(Synchronous), and Syn-GRPO~(Asynchronous).}
\vspace{-1em}
\label{tab:training_time}
\end{table}